\documentclass[journal]{IEEEtran}
\ifCLASSINFOpdf
\else
\fi

\usepackage{times}
\usepackage{epsfig}
\usepackage{graphicx}
\usepackage{amsmath}
\usepackage{amssymb}
\usepackage{xcolor}
\usepackage[utf8]{inputenc}

\DeclareUnicodeCharacter{2212}{-}

\usepackage{caption}
\usepackage{subcaption}
\usepackage{float}
\usepackage{multirow}
\usepackage{algorithm}
\usepackage{algpseudocode}
\usepackage[pagebackref=true,breaklinks=true,letterpaper=true,colorlinks,bookmarks=false]{hyperref}
\usepackage[capitalise]{cleveref}

\newcommand{\Loss}[2]{\mathcal{L}_{#1}^{#2}}

\newcommand{\revision}[1]{#1}
\newcommand{\camera}[1]{#1}

\newcommand{\E}[1]{\mathcal{E}_{#1}}

\newcommand{\C}[1]{\mathcal{C}_{#1}}

\begin{document}
\title{Adversarial Learning of Disentangled and Generalizable Representations of Visual Attributes}

\author{James~Oldfield,
        Yannis~Panagakis,
        and~Mihalis~A.~Nicolaou
\thanks{J. Oldfield is with the Department of Electronic Engineering and Computer Science, Queen Mary University of London, UK.}
\thanks{Y. Panagakis is with the Department of Informatics and Telecommunications, University of Athens, Greece.}
\thanks{M. A. Nicolaou is with the Computation-based Science and Technology Center at the Cyprus Institute.}
}

\markboth{Journal of \LaTeX\ Class Files,~Vol.~X, No.~X, 2019}%
{Shell \MakeLowercase{\textit{et al.}}: Bare Demo of IEEEtran.cls for IEEE Journals}

\maketitle

\begin{abstract}
Recently, a multitude of methods for {image-to-image} translation have demonstrated impressive results on problems such as multi-domain or multi-attribute  transfer.  The vast majority of such works leverages the strengths of adversarial learning and deep convolutional autoencoders to achieve realistic results by well-capturing the target data distribution.  Nevertheless, the most prominent representatives of this class of methods do not facilitate semantic structure in the latent space, and usually rely on binary domain labels for test-time transfer.  This leads to rigid models, unable to capture the variance of each domain label. In this light,  we propose a novel adversarial learning method that (i) facilitates the emergence of latent structure by semantically disentangling sources of variation, and (ii) encourages learning generalizable, continuous, and transferable latent codes that enable flexible attribute mixing.  This is achieved by introducing a novel loss function that encourages representations to result in uniformly distributed class posteriors for disentangled attributes.  In tandem with an algorithm for inducing generalizable properties, the resulting representations can be utilized for a variety of tasks such as intensity-preserving multi-attribute image translation and synthesis, without requiring labelled test data.   We demonstrate the merits of the proposed method by a set of qualitative and quantitative experiments on popular databases such as MultiPIE, RaFD, and BU-3DFE, where our method outperforms other, state-of-the-art methods in tasks such as intensity-preserving multi-attribute transfer and synthesis.
\end{abstract}

\IEEEpeerreviewmaketitle

\section{Introduction}
\label{sec:intro}
 
\looseness-1
Recently, deep generative models trained through adversarial learning \cite{goodfellow2014generative} (GANs) have been shown capable of generating naturalistic visual data that appear  authentic to human observers.   GAN-based generative models have found an immensely wide and diverse range of applications, ranging from vision-related tasks such as \revision{image synthesis and} image-to-image translation \cite{cycleGAN,pix2pix,starGAN} \revision{\cite{Shaham_2019_SinGAN}}, style transfer, \cite{cycleGAN}, medical imaging~\cite{low-dose-ct,medical-synth},  and artwork synthesis~\cite{art-gan}, to scientific applications in astronomy and physics \cite{mustafa2019cosmogan}, to mention but a few examples.  

\looseness-1One of the most prominent applications of GANs lies in Image-to-Image Translation (IIT).  
Specifically,
IIT methods \cite{cycleGAN,starGAN,pix2pix,discoGAN} learn a non-linear mapping of an image in a source domain to its corresponding image in a target domain. The notion of domain varies, depending on the application. For instance, in the context of super-resolution, the source domain consists of low-resolution images while the corresponding high-resolution images belong to the target domain \cite{ledig2017photo,wang2018esrgan}. In a visual attribute transfer setting, `domain' denotes facial images with the same attribute that describes either intrinsic facial characteristics  (e.g., identity, facial expressions, age, gender) or captures external sources of appearance variation related, for example, to different poses and illumination conditions. In this context, the task usually lies in  synthesizing a novel image, that changes  visual attributes to match a specific value.

\begin{figure*}[h]
    \centering
     \includegraphics[width=1.0\linewidth]{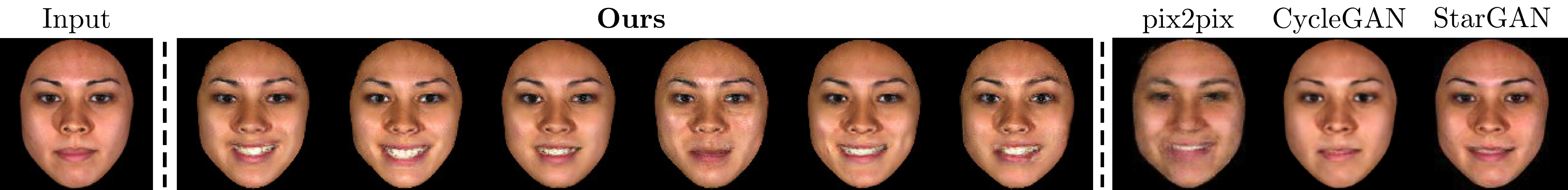}
    \caption{\looseness-1Expression transfer results with the same  target expression (smile). The proposed method excels at preserving intra-attribute variance, and can readily generate diverse synthetic images with varying expression intensity.
}
    \label{fig:intensity-preserving}
\end{figure*}

Deep generative models for image-to-image translation implement a mapping between two \cite{cycleGAN} or multiple \cite{starGAN} image domains, in a  paired \cite{pix2pix} or unpaired \cite{cycleGAN,starGAN,unit,discoGAN,MUNIT} fashion. Despite their merits in pushing forward the state of the art in image generation, we posit that the widely adopted image-to-image translation models (namely the CycleGAN \cite{cycleGAN}, Pix2Pix \cite{pix2pix} and StarGAN \cite{starGAN})  also come with a set of shortcomings. For instance, none of these methods impose any model constraints that can result in semantically meaningful latent structure, sacrificing model flexibility.  In addition, the generated images do not cover the variance of the target domain; in fact, in most cases a single image is generated given a target binary attribute value, which is also required at test time.  For example, changing the ``smile'' attribute of a facial image will always lead to a smile of specific intensity--as shown in \cref{fig:intensity-preserving}.  This also inhibits the image generation by transfer of unseen or unlabelled attributes that do not appear in the given training set.

\looseness-1In this paper, we propose a novel method\footnote{Code for reproducing our results can be found at: \href{https://github.com/james-oldfield/adv-attribute-disentanglement}{https://github.com/james-oldfield/adv-attribute-disentanglement}} that facilitates learning disentangled, generalizable, and continuous representations of visual data with respect to attributes acting as sources of variation.  In contrast to current popular approaches, the proposed method can readily be used to generate images that contain varying intensity expressions (\cref{fig:intensity-preserving})  in images,  while also being equipped with several other features that enable the flexible generation of novel hybrid imagery on unseen data.   In more detail, the key contributions of this work are summarized below.

\begin{figure*}
    \centering
    \includegraphics[width=1.0\linewidth]{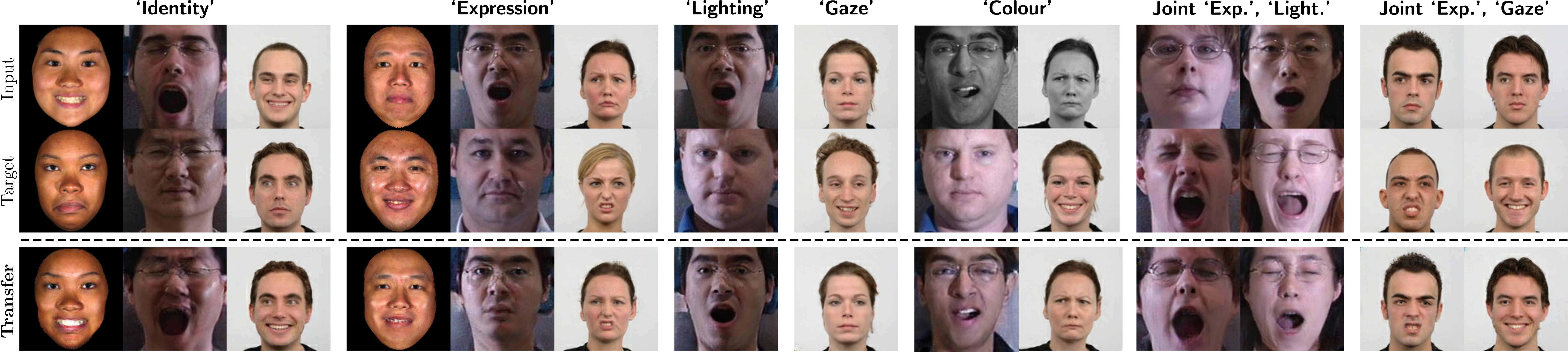}
    \caption{Multiple-attribute transfer across  databases and attributes. \textbf{Row 1}: Input image; \textbf{Row 2}: Target image; \textbf{Row 3}: Single or Joint attribute transfer (zoom in for better quality).}
    \label{fig:all-transfers}
\end{figure*}

\begin{itemize}
\item \looseness-1Firstly, a novel loss function for learning disentangled representations is proposed.  The loss function ensures that latent representations corresponding to an attribute (a) have {\it discriminative} power within the attribute class, while (b) being {\it invariant} to sources of variation linked to other attributes.  For example, given a facial image with a specific identity and expression, representations of the {\it identity} attribute should classify all identities well, but should fail to classify the expressions well: that is, the conditional distribution of the class posteriors should be {\it uniform} over the expression labels.

\item Secondly, we propose a novel approach that encourages the disentangled representations \revision{of {\it multiple} attributes} to be {\it generalizable}.
The particular loss function enables the network to generate realistic images that are classified appropriately, even when sampling representations from different images. This enables the representations to well-capture the attribute variation as shown in~\cref{fig:intensity-preserving}, in contrast to other methods that simply consider a target label for transfer.

\item \looseness-1Finally, we provide a set of rigorous experiments to demonstrate the capabilities of the proposed method on  databases such as MultiPIE, BU-3DFE, and RaFD.  Given generalizable and disentangled representations on a multitude of attributes (e.g., expression, identity, illumination, gaze, color), the proposed method can perform {\it arbitrary} combinations of the latent codes in order to generate novel imagery.  For example, we can swap an arbitrary number of attribute representations amongst test samples to perform intensity-preserving multiple attribute transfer and synthesis, {\it without} knowing the test labels.  What's more, we can combine an arbitrary number of latent codes in order to generate novel images that preserve a mixture of characteristics--a particularly useful feature for tasks such as data augmentation.  Both qualitative and quantitative results corroborate the improved performance of the proposed approach over state-of-the-art image-to-image translation methods.

\end{itemize}

\section{Methodology}
\label{sec:meth}
\looseness-1In this section, we provide a detailed description of the methodology proposed in this work, which focuses on learning disentangled and generalizable latent representations of visual attributes. Concretely, in~\cref{sec:meth:genmodel} we describe the generative model employed in this work.  In~\cref{sec:meth:disentangle}, we introduce the proposed loss functions tailored towards disentangling representations in latent space, such that they (i) well-capture variations that relate to a given attribute (e.g., {\it identity}) by enriching features with discriminative power, and (ii) {\it fail} to classify any other attributes well, by encouraging the classifier posterior distribution over values of other attributes (e.g., {\it expression}) to be uniform. Furthermore, in~\cref{sec:meth:generalise}, we describe an optimization procedure  tailored towards encouraging recovered representations to be {\it generalizable}, that is, can be utilized towards generating novel, realistic images from arbitrary samples and attributes. The full objective employed is described in~\cref{sec:fullobjective} by incorporating the above ideas in an adversarial learning framework, while  implementation details regarding the full network that is trained in an end-to-end fashion are provided in~\cref{sec:meth:implementation}.  Finally, an overview of the proposed method is illustrated in~\cref{fig:overview}.

\subsection{Generative Model}
\label{sec:meth:genmodel}
\looseness-1Given a dataset with $N$ samples, we assume that each sample $\mathbf{x}^{(i)}$ is associated with a set of $M$ attributes that act as sources of visual variation  (such as identity, expression, lighting).  If not omitted, the superscript $i$ denotes that $\mathbf{x}^{(i)}$ is the $i$-th sample in the dataset or mini-batch. We further assume a set of labels $y_m(\mathbf{x})$ corresponding to each attribute $m$.    We aim to recover disentangled, latent representations $\mathbf{z}_m$ that capture variation relating to attribute $m$, while being invariant to variability arising from the remaining $M-1$ attributes.  We  assume the following generative model,
\begin{equation}
\label{eq:generative}
\mathbf{\tilde{x}}=\mathcal{D}(\mathcal{E}(\mathbf{x}))=\mathcal{D}(\mathbf{z})=\mathcal{D}(\mathbf{z}_0, \dots, \mathbf{z}_M)
\end{equation}
where $\mathbf{z}_m=\mathcal{E}_m(\mathbf{x})$ is an encoder mapping the sample $\mathbf{x}$ to a space that preserves variance arising only from attribute $m$, while $\mathcal{D}$ is a decoder mapping the representations back to input space.  Note that when $m=0$, the corresponding representations $\mathbf{z}_0$ represent variation present in an image that does not relate to any of the specified attributes.  For example, assuming a dataset of facial images, if $M=2$ with $\mathbf{z}_1$ corresponding to {\it identity} and $\mathbf{z}_2$ to {\it expression}, $\mathbf{z}_0$ captures other variation.  To ensure that the set of $M+1$ representations faithfully reconstruct the original image $\mathbf{x}$, we impose a standard reconstruction loss,
\begin{equation}
\label{eq:recon}
\mathcal{L}_{rec}=\mathbb{E}_{\mathbf{x}}\Big[
    || \mathbf{x} - \mathcal{D}(\mathcal{E}(\mathbf{x})) ||_1
\Big].
\end{equation}

\subsection{Learning Disentangled Representations} 
\label{sec:meth:disentangle}
\looseness-1Our aim is to define a transformation that is able to generate disentangled representations for specific attributes that act as sources of variation, while being invariant with respect to other variations present in our data.  To this end, we introduce a method for training the encoders $\mathcal{E}_m$ arising in our generative model (Eq. \ref{eq:generative}), where  resulting representations $\mathbf{z}_m$ have discriminative power over attribute $m$, while yielding maximum entropy class posteriors for each of the other $M-1$ attributes.  In effect, this prevents any contents relating to other specified attributes besides $m$ from arising in the resulting representations.  To tackle this problem, we propose a composite loss function as discussed below.

{\bf Classification Loss.}  We firstly employ a loss function to ensure that the obtained representations $\mathbf{z}_m = \mathcal{E}(\mathbf{x})$ well-classify variation related to attribute $m$.  This is done by feeding the representations directly into the penultimate fully-connected layers of a classifier $\C{m}$, and minimizing the negative log-likelihood of the ground truth labels given a latent encoding, 
\begin{align}
\label{eq:z-cls}
\mathcal{L}^{\mathbf{x}}_{cls}=&\mathbb{E}_{\mathbf{x}}\bigg[
\frac{1}{M}
\sum_{m=1}^M\bigg(
    -\log\mathcal{C}_{m}\Big(
        y_{m}(\mathbf{x}) \mid
        \mathcal{E}_{m}(\mathbf{x})
    \Big)
      \bigg)\bigg].
\end{align}
Minimizing ~\cref{eq:z-cls} ensures that the latent representations must retain a desirable amount of related high-level information in order to induce the correct classification.

{\bf Disentanglement Loss.}  \looseness-1Classification losses, as defined above, ensure that the learned transformations leads to representations that are enriched with information related with the particular ground-truth label, and have been  employed in different forms in other works such as \cite{starGAN}.  However, as we demonstrate experimentally in \cref{sec:modelexploration}, it is not  reasonable to expect that the classification loss alone is sufficient to disentangle the latent representations $\mathbf{z}_m$ from other sources of variations arising from the other $M-1$ attributes.  Hence, to encourage disentanglement, we impose an additional  loss on the conditional label distributions of the classifiers, given the corresponding representations.  In more detail, we posit that the class posterior for each attribute $m$ given the latent representations $\mathbf{z}_m$ induced by the encoder $\mathcal{E}_m(\mathbf{x})$ for every other distinct  attribute $m'$ should be a {\it uniform} distribution. 
We impose this soft-constraint by minimizing the cross entropy loss between a uniform distribution and the classifier class posteriors, 
\begin{align}
\label{eq:ambiguity-objective}
\mathcal{L}_{dis}^m=&\mathbb{E}_{\mathbf{x}}\bigg[
      -\frac{1}{(M-1)}
      \mathop{\sum_{m'=1}^{M}}_{m' \neq m}
      \sum_{i=1}^{|m'|}
      \frac{1}{|m'|} \log \C{m'}\Big(y_{m'} \mid \mathcal{E}_m(\mathbf{x})\Big)_i
      \bigg],
\end{align}
\looseness-1where $m'$ iterates over all other attributes and  $|m'|$ is the number of classes for attribute $m'$. In other words, we impose that each encoder $\mathcal{E}_m$ must map to a representation that is correctly classified with respect to~\textit{only} the relevant attribute $m$ (with \cref{eq:z-cls}), and that the representation is such that the conditional label distribution given its mapping, for every other attribute, has maximum entropy (with \cref{eq:ambiguity-objective}). In effect, the proposed disentanglement loss function {\it filters out} information related to variation arising from other specified attributes besides the attribute of interest.  We note that the final loss function averages over all attributes, that is $\mathcal{L}_{dis}=\frac{1}{M+1}\sum_{m=0}^M\mathcal{L}_{dis}^{m}$.  In particular, for $m=0$ we ensure that the representations obtained via $\mathbf{z}_0=\mathcal{E}_0(\mathbf{x})$ are invariant to {\it all} $M$ attribute variations, while capturing only variations that are not related to any of the $M$ attributes.

\begin{figure*}[h]
    \centering
    \includegraphics[scale=0.38,trim={0.5cm 0.2cm 0.5cm 0.2cm},clip]{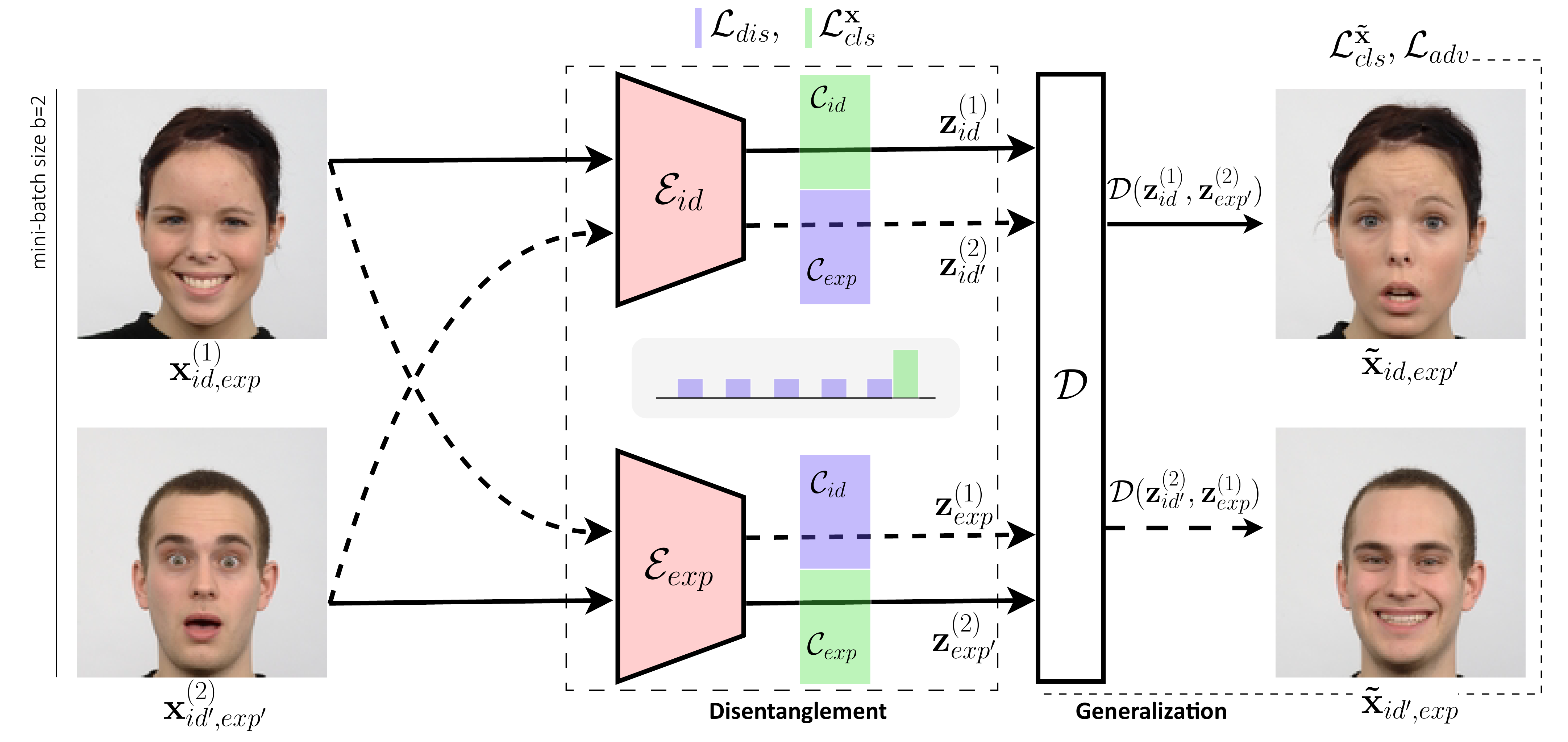}
    \caption{\looseness-1Overview of the proposed method, in a simple setting with two samples and two attributes: {\it identity} and {\it expression}.  Samples are mapped to distinct disentangled representations $\mathbf{z}^{(i)}_m$ by utilizing the classification $\mathcal{L}_{cls}^{\mathbf{x}}$ and disentanglement losses $\mathcal{L}_{dis}$, encouraging the representations to well-classify identity (expression), and the classifier posteriors to be uniform over the expression (identity) labels.  To enable the decoder $\mathcal{D}$ to generate novel images at test time, generalization is encouraged by permuting the latent representations for each attribute sample-wise.  Utilizing a classification $\mathcal{L}_{cls}^{\mathbf{\tilde{x}}}$ and adversarial loss $\mathcal{L}_{adv}$, the synthesized images $\mathbf{\tilde{x}}$ are encouraged to be both realistic, as well as be classified correctly according to the given attribute.  Our method does not require the synthesized images to exist in the dataset.  {Note: For clarity of presentation, we omit $\mathcal{E}_0$, the encoder handling variations unrelated to attributes $M$. \label{fig:overview}}
    }
    \label{fig:my_label}
\end{figure*}

\subsection{Learning Generalisable Representations}
\label{sec:meth:generalise}
\looseness-1The loss function described in~\cref{sec:meth:disentangle} encourages the representations $\mathbf{z}_m$ generated by the corresponding encoders to capture the variation induced by attribute $m$, while being invariant to variation arising from other sources.  In this section, we provide a simple, effective method for ensuring that the derived representations are {\it generalizable} over unseen data (e.g., new identities or expressions in facial images that are  unseen during training), while at the same time yielding the expected semantics in the generated images.   Similarly to the previous section, we utilize classifier distributions to learn generalizable representations {\it without} requiring the ground-truth pair for any combination of labels.  

We assume a mini-batch of size $b$.  During each forward pass, we randomly shuffle the representations for each attribute along the batch dimension, which when passed through the decoder provides a new synthesized sample, $\mathbf{\tilde{x}'}$,
\begin{equation}
\label{eq:shuffle-procedure}
\mathbf{\tilde{x}}'=\mathcal{D}(\mathbf{z}')=\mathcal{D}(\mathbf{z}_0,\mathbf{z}_1^{(r_1)},\dots,\mathbf{z}_M^{(r_M)}),
\end{equation}
\looseness-1where $r_1,\dots,r_M\in[1,b]$ are random integers indexing the mini-batch data (Fig. \ref{fig:overview}).   In essence, this leads to a synthesized sample $\mathbf{\tilde{x}}'$.  Since we know the ground-truth label values that the attributes should be taking in the synthesized sample $\mathbf{\tilde{x}}'$, we can enforce a classification loss on $\mathbf{\tilde{x}}'$ by minimizing the negative log-likelihood of the expected classes for each attribute,
\begin{equation}
\label{eq:transfer-cls}
\mathcal{L}_{cls}^{\mathbf{\tilde{x}}}=\mathbb{E}_{\mathbf{\tilde{x}}'}\left[
    \frac{1}{M}
    \sum_{m=1}^M
    -\log\mathcal{C}_{m}\left(
        y_{m}(\mathbf{\tilde{x}'}) \mid
        \mathbf{\tilde{x}}'
    \right)
\right].
\end{equation}
Note that this process is further illustrated in~\cref{fig:overview}. Finally, we highlight that the above loss is at an advantage over~\textit{paired} methods (such as~\cite{pix2pix}), in that no direct access to the corresponding target $\mathbf{x}'$ is required, since no reconstruction loss is imposed {on $\mathbf{\tilde{x}}'$}.

{\bf Adversarial Loss} In order to induce adversarial learning in the proposed model and encourage generated images to match the data distribution, we further impose an adversarial loss in  tandem with  (\ref{eq:transfer-cls}), 
\begin{equation}\label{eq:gan-objective}
  \Loss{adv}{}=\mathbb{E}_{\mathbf{x}}[\log D(\mathbf{x})] + \mathbb{E}_{\mathbf{\tilde{x}'}}[\log(1-D(\mathbf{\tilde{x}'}))]
\end{equation}
This ensures that even when  representations $\mathbf{z}_m$ are shuffled across data points, the synthesized sample will both (i) be classified according to the sample/embedding combination (Eq. \ref{eq:gan-objective}), as well as (ii) constitute a realistic image. 

\subsection{Full Objective}
\label{sec:fullobjective}
The proposed method is trained end-to-end, using the full objective as grouped by the set of variables we are optimizing for
\begin{align}\label{eq:full-objective}
\Loss{G}{}= & \Loss{adv}{} + \Loss{dis}{} + \Loss{cls}{\mathbf{x}} + \Loss{cls}{\mathbf{\tilde{x}}} + \Loss{rec}{} \nonumber\\
\Loss{D}{}= & -\Loss{adv}{},\;\;\; \Loss{C}{}=  \Loss{cls}{c},
\end{align}
where $\Loss{D}{}$ is the combined loss for the discriminator, $\Loss{C}{}$ is for the classifiers, and $\Loss{G}{}$ is for the encoders and decoder. We control the relative importance of each loss term $\Loss{i}{}$ with a corresponding $\lambda_i$ hyperparameter.

\subsection{Implementation}
\label{sec:meth:implementation}
At train-time, we sample image mini-batches  of size $b=16$ randomly. In each iteration, we shuffle each of the $M$ attribute encodings along the batch dimension before concatenating depth-wise, and feeding into the decoder (i.e. \cref{eq:shuffle-procedure}), to encourage the network to be able to flexibly pair any combination of values of the attributes.

{\bf Network Architecture.} We define $M+1$ instances of the encoders (one for each of the $M$ explicitly modeled attributes, and an additional encoder to capture the remaining sources of variation). Each encoder $\E{m}$ is a separate convolutional encoder based on the first half of~\cite{neural-style-transfer} up to the bottleneck. The decoder $\mathcal{D}$ depth-concatenates all $M+1$ latent encodings and then upsamples via~\cite{resize-conv} to reconstruct the input image. The specific network architecture choices made are presented in Table~\ref{table:generator}.  We adopt the deeper PatchGAN variant proposed in~\cite{starGAN} for the discriminator, that classifies overlapping image patches rather than the entire image, thus encouraged to preserve high-frequency information. The $M$ classifiers $\C{m}$ are simple shallow CNNs--trained on the images in the training set to correctly classify the labels of its designated attribute $m$--with a final dense layer that outputs the logits for the classes of the appropriate attribute $m$. The classifier internals are detailed in \cref{table:classifier}. Here we highlight our design choice to make the output of the final layer of the encoders \verb+enc-3+ the same dimensionality as the required input to the penultimate layer of the classifiers, \verb+class-logits+, in order to instill class-specific representations directly on the latent attribute representations by feeding in these latent representations directly to this point in the classifier.

\begin{table*}[h]
\resizebox{\textwidth}{!}{
\begin{tabular}{|l|l|l|l|l|}
\hline
Section & Name & Dimensions (In) & Dimensions (Out) & Layers \\ \hline
\multirow{3}{*}{Encoder $m$} & enc-1 & $(b, 128, 128, 3)$ & $(b, 128, 128, 32)$ & r\_pad(3) $\rightarrow$ Conv2d(f=32, k=7, s=1) $\rightarrow$ IN $\rightarrow$ ReLu \\ \cline{2-5} 
 & enc-2 & $(b, 128, 128, 32)$ & $(b, 64, 64, 64)$ & Conv2d(f=64, k=3 s=2) $\rightarrow$ IN $\rightarrow$ ReLu \\ \cline{2-5} 
 & enc-3 & $(b, 64, 64, 64)$ & $(b, 32, 32, 128)$ & Conv2d(f=128, k=3 s=2) $\rightarrow$ IN $\rightarrow$ ReLu \\ \hline
\multirow{4}{*}{Decoder} & dec-res-\{1..6\} & $(b, 32, 32, 128 * (M+1))$ & $(b, 32, 32, 128 * (M+1))$ & Conv2d(f=128*($M+1$), k=3 s=1) $\rightarrow$ IN $\rightarrow$ ReLu \\ \cline{2-5} 
 & dec-1 & $(b, 32, 32, 128 * (M+1))$ & $(b, 64, 64, 64)$ & r\_pad(1) $\rightarrow$ resize(128) $\rightarrow$ Conv2d(f=64, k=3, s=2) $\rightarrow$ IN $\rightarrow$ ReLu \\ \cline{2-5} 
 & dec-2 & $(b, 64, 64, 64)$ & $(b, 128, 128, 32)$ & r\_pad(1) $\rightarrow$ resize(256) $\rightarrow$ Conv2d(f=32, k=3, s=2) $\rightarrow$ IN $\rightarrow$ ReLu \\ \cline{2-5} 
 & dec-3 & $(b, 128, 128, 32)$ & $(b, 128, 128, 3)$ & r\_pad(3) $\rightarrow$ Conv2d(f=3, k=7, s=1) $\rightarrow$ tanh \\ \hline
\end{tabular}%
}
\caption{\revision{Encoder and Decoder architecture: $b$ denotes the mini-batch size and $M$ the total number of specified attributes. `IN' refers to the Instance Normalization operation \cite{instance-norm}. The output of \texttt{enc-3} is taken as the latent representation $\mathbf{z}_m$ for encoder $m$.}}
\label{table:generator}
\end{table*}
\begin{table*}
\centering
\begin{tabular}{|l|l|l|l|l|}
\hline
Section & Name & Dimensions (In) & Dimensions (Out) & Layers \\ \hline
\multirow{3}{*}{Classifiers} & block-1 & $(b, 128, 128, 3)$ & $(b, 64, 64, 64)$ & pad(1) $\rightarrow$ Conv2d(f=64, k=4, s=2) $\rightarrow$ LeakyRelu \\ \cline{2-5} 
 & block-2 & $(b, 64, 64, 64)$ & $(b, 32, 32, 128)$ & pad(1) $\rightarrow$ Conv2d(f=128, k=4, s=2) $\rightarrow$ LeakyRelu \\ \cline{2-5} 
 & class-logits & $(b, 32, 32, 128)$ & $(b, |m|)$ & Flatten $\rightarrow$ Dense($|m|$) \\ \hline
\end{tabular}%
\caption{\revision{Classifier architecture: $b$ again denotes the mini-batch size and $|m|$ the total number of classes for attribute $m$.}}
\label{table:classifier}
\end{table*}

\section{Related Work}
\label{sec:related-work}
\revision{In this section, we first provide an overview of recent image-to-image translation methods for various problem settings including: paired, unpaired, and multi-domain (\cref{sec:related:imagetranslation}). Subsequently, we discuss related work for obtaining generalizable and disentangled representations, and highlight the novelty of our approach (\cref{sec:related:generalizable}).}

\subsection{Image-to-image Translation}
\label{sec:related:imagetranslation}

\revision{Image-to-image translation (IIT) employing deep neural networks and/or adversarial training has enjoyed widespread success across a number of computer vision tasks \cite{pix2pix,cross-domain-seg,texture-gan,dance-now},
with GAN-based approaches being capable of producing sharp synthetic images. Another popular approach for image synthesis is VAE-based methods \cite{kingma2014autoencoding}. VAEs however are often prone to generating blurry images \cite{zhao2017vae_understanding}, and thus many recent variants have been proposed to alleviate this, including the use of hierarchical encoders \cite{razavi2019generating} or by combining VAEs with GANs \cite{huang2018introvae}}.
At a high level, the goal of IIT translation can be seen as learning the mapping between two (or more) sets of images, with the assumption that the images in each set share some kind of visual characteristic or attribute (e.g. a painting style, a particular hair colour, etc.).

The seminal \textbf{pix2pix} \cite{pix2pix} paper introduced a supervised approach to the task of IIT using convolutional autoencoders and a conditional adversarial loss \cite{mirza2014conditional}, but makes the restrictive assumption of \textit{paired} correspondence between the training data points in each image set. Inspired by these shortcomings, \textbf{CycleGAN} \cite{cycleGAN} proposed the so-called ``cycle-consistency'' loss term, which ultimately led to the capability to handle training images that are \textit{unpaired} between image domains. Multiple other works \cite{discoGAN,dualGAN} also circumvented such limitations with similar additional loss terms, including the recent \textbf{StarGAN} \cite{starGAN}, which also achieves state-of-the-art performance across multiple image domains. It is for this reason that we benchmark our experiments primarily against \textbf{StarGAN}, as it shares the largest subset of the many capabilities that our approach affords. Importantly however, none of these methods are designed to facilitate a latent attribute space in the manner that ours is, and are therefore limited to performing IIT by conditioning on a small set of known discrete label expression values, and hence do not permit continuous-valued attribute transfer, or label-free test-time attribute transfer. Due to the attributes being represented in this discrete manner, the variance within the attributes themselves is also lost, in contrast to our proposed method.

The recent \textbf{StyleGAN} \cite{styleGAN} achieves high resolution results and SOTA unconditional image generation via techniques such as progressive growing \cite{karras2017progressive} and multi-resolution style modulation via AdaIN \cite{adain}. However, no such IIT capabilities are present, and such a generator network with attribute modulation via AdaIN could in theory be incorporated into our own method. Using similar style modulation techniques for the purposes of IIT, \textbf{MUNIT} \cite{MUNIT} decomposes images into a style and content code, and in contrast to other image translation works, modulates the `style' content explicitly via AdaIN. Under our framework, we can view the `style' of image as being analogous to a single attribute (such as `expression'). 

Taking a similar approach to IIT by learning invariant image factors, {\bf FaderNets}~\cite{fader} adopt the approach of learning an image attribute representation (i.e. a person's identity) that is invariant to any other variation from attributes such as ``glasses'', and ``facial hair''.   However, FaderNets do not facilitate the semantic decomposition of the latent space--instead, for the purposes of IIT, FaderNets rely on a binary value that corresponds to the target attribute value.  This means that, in contrast to the proposed method, labels are required at test-time. This highlights a unique feature of the proposed method, that can afford intensity-preserving, continuous-valued multi-attribute transfer \revision{of the particular patterns of a target image of choice. Concurrent work \cite{Lin_2019_RelGAN} proposes a method for continuous-valued attribute transfer by manually specifying the relative difference between the input image and target image's attribute values with a real-valued scalar. Our method in contrast offers an additional flexibility--we condition on an entire target face image, from which a multi-dimensional target attribute value is {\it automatically} extracted. Using this high-dimensional target attribute representation  provides much more fine-grained control over  attribute transfer compared to when using a single target value. For example, the particular countenance of a target image can be captured in the resulting representations, while specifying this manually with a single real number--as in {\it RelGAN}--can be much more challenging.}
Finally, it is important to note that the problem formulation of FaderNets trades-off disentanglement (invariance) with reconstruction quality, unlike our method that utilizes a specific latent semantic decomposition where improving disentanglement does not have a negative impact on reconstruction quality.

\subsection{Generalizable \camera{and Disentangled} Representations}
\label{sec:related:generalizable}

\revision{
A popular approach to encouraging representations to be generalizable is to employ a training scheme whereby the latent contents are `shuffled' in some manner to produce a new synthetic image with the expected combination of visual attributes.
For example, \cite{Bao_2018_idPres} perform identity transfer by disentangling the `identity' attribute from the remaining modes of variation, using such a technique. In a similar manner, \cite{Jha_2018_cycleVAE} swaps the latent variable between images to encourage the learnt representations to be generalizable, simultaneously disentangling a target attribute from all the unspecified attributes. These approaches however are limited to the case of disentangling only a single attribute. Specifically, the remaining attributes aside from the single target attribute remain entangled, and thus such models are limited to single-attribute image translation.
Motivated by this, our method proposes a multiple attribute extension to these approaches, in tandem with the disentanglement loss. Such a feature of our model thus allows one to jointly transfer the semantic image contents pertaining to multiple attributes at a time.

\camera{The idea of minimizing the cross entropy between a uniform distribution and a classifier's class posteriors has been explored for tasks such as synthesizing novel artwork \cite{can} and for learning domain-invariant representations \cite{domaininv}.
An important distinction between the latter and our work however is that our proposed approach utilizes the raw pre-trained image classifiers with no additional parameters or training needed.}}

\section{Experiments}
\label{sec:experiments}
\looseness-1 In this section, we present a set of rigorous qualitative and quantitative experiments on real-world datasets, in order to validate the properties of the proposed method, and demonstrate its merits on multiple real-world datasets.  

\looseness-1Concretely, we experiment on datasets such as MultiPIE, BU-3DFE, and RaFD, with experimental settings discussed in more detail in~\cref{sec:datasets}.   Subsequently, the proposed method is utilized towards learning {\it disentangled} and {\it generalizable} representations on various categorical attributes, including {\it identity}, {\it expression}, {\it illumination}, {\it gaze}, and even {\it color}.  In more detail, in~\cref{sec:modelexploration} we present a set of ablation studies that validate the disentangled nature of resulting embeddings, both qualitatively by visualizing t-SNE embeddings, as well as quantitatively by evaluating classifier predictive distribution and entropy. In~\cref{sec:experiments:synthesis}, we detail experiments related to expression synthesis in comparison to SOTA image-to-image translation models, covering the entire span of datasets that are under consideration.  Subsequently, in~\cref{sec:experiments:atttransfer}, results on intensity-preserving transfer of arbitrary attributes are presented, with the proposed representations well-capturing attribute variation.  In the same section, the joint transfer of multiple attributes is also demonstrated.  Subsequently, in~\cref{sec:experiments:novelidinter}, we further evince the generalizable nature of the latent representations for each attribute. By performing weighted combinations of the represenations over multiple samples, we can generate novel unseen data, as for example novel identities.  To further demonstrate the continuous properties of the derived representations, along with generalization and disentanglement, we perform latent space interpolations between images with arbitrary attribute values.

\subsection{Datasets and Experimental Setting}
\label{sec:datasets}
For all experiments we follow \cite{starGAN} and adopt the Wasserstein-GP \cite{wgan,wgan-improved} GAN objective,
\begin{equation}
\mathcal{L}_{adv}=\mathbb{E}_\mathbf{x}[ D(\mathbf{x}) ] - \mathbb{E}_{{\mathbf{\tilde{x}}'}}[ D({\mathbf{\tilde{x}}'}) ] - \lambda_{gp}\mathbb{E}_{\mathbf{\hat{x}}}[ (|| \nabla_{\mathbf{\hat{x}}} D(\mathbf{\hat{x}}) ||_2-1)^2 ],
\end{equation}
with $\mathbf{\tilde{x}}'$ denoting a synthesised image as defined in \cref{eq:shuffle-procedure}, and $\mathbf{\hat{x}}$ denoting an image sampled uniformly along a straight line between pairs of the real and synthesised imagery. We set the both the gradient penalty and reconstruction weights as $\lambda_{rec}:=10, \lambda_{gp}:=10$.  In what follows, we present a brief description of the datasets employed in the experimental section, along with details on training and testing set selection, noting that same dataset splits are used with  all evaluated models to ensure fair comparisons across datasets.\\
\\
{\bf  MultiPIE.} The CMU Multi Pose Illumination and Expression (MultiPIE)~\cite{multi-pie} database consists of over $750,000$ images that include $337$ subjects captured under a challenging range of variation over four different sessions.  For each subject, images span $15$ different poses, $19$ illumination conditions, as well as $6$ different facial expressions.  For our experiments, we use the forward-facing subset of MultiPIE, jointly modelling attributes `identity', `expression' and `lighting'.  The training set consists of $686$ images for each of the emotion classes (`neutral', `scream', `squint', and `surprise'). The first 10 identities are held out and comprise the test set. \\
\\
{\bf BU-3DFE (BU).} The Binghampton University Facial Expression Database (BU)~\cite{BU} includes data captured from $100$ subjects, covering a wide range of age, gender, race, and expression variation (with both prototypic expressions and 4-intensity levels).  The database amounts to over $2500$ texture images.  We utilize the entire set of 2D frontal projections for training, with 2160 images used for training (90 different identities along with varying intensity expressions), reserving the $240$ remaining images for the test set (including $10$ different identities and corresponding expressions).\\
\\
{\bf RaFD.} The Radboud Faces Database (RaFD) database~\cite{rafd} consists of images of $67$ subjects, with variance in terms of gender, age, and ethnicity.  Each subject is recorded displaying $8$ different emotional expressions, $3$ different gazes, and $3$ poses, leading to $4824$ high-quality images in total.  For our experiments we utilize the front-facing poses, holding out the (numerically) first $8$ identities for the test set (for a total of 192 test set images), while we use the remaining $1416$ images for training.

\begin{figure*}[h]
    \centering
    \begin{subfigure}{.45\linewidth}
        \centering
        \includegraphics[width=0.95\linewidth]{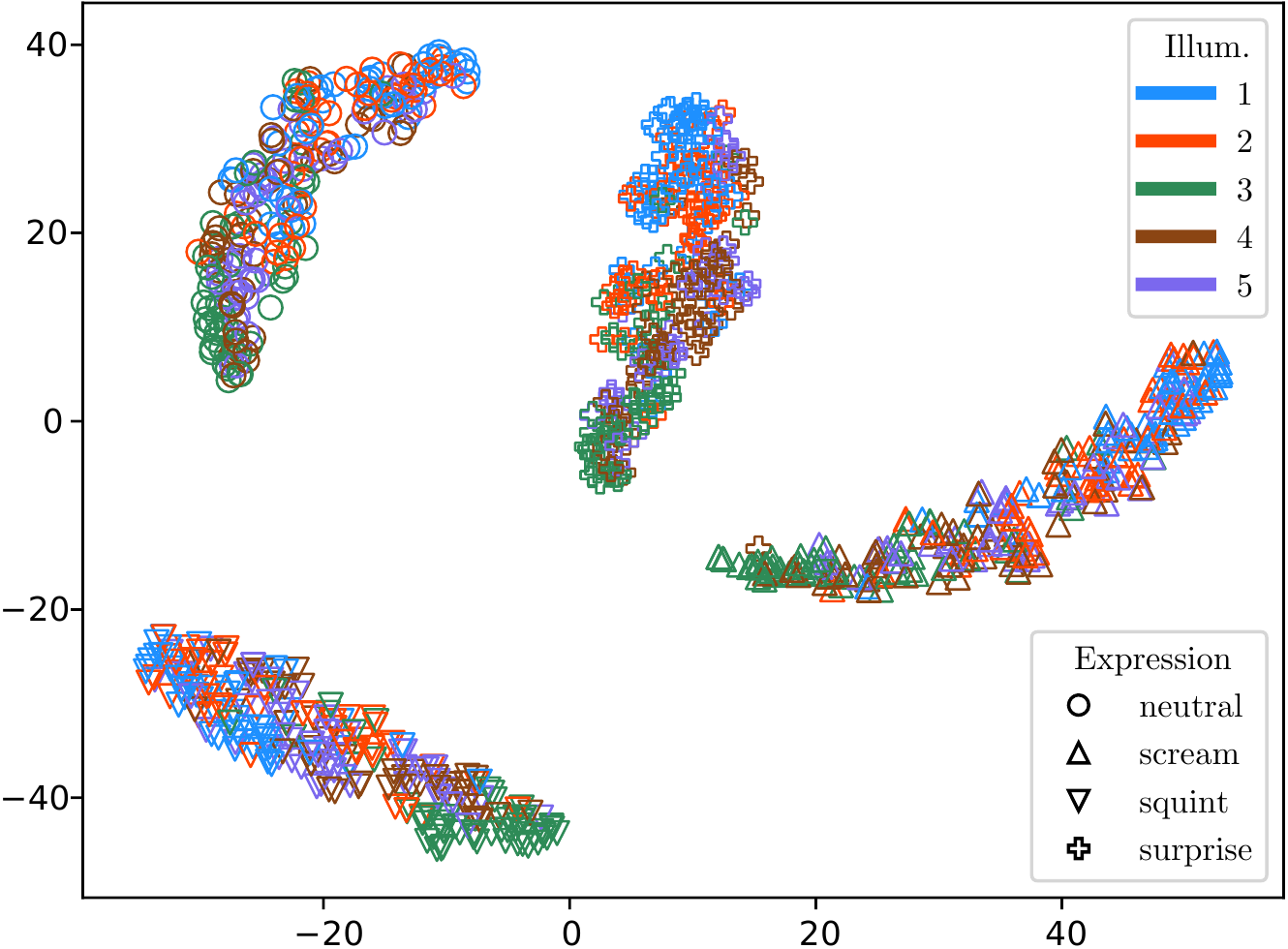}
        \caption{}
        \label{fig:dim-without}
        \qquad
    \end{subfigure}
    \begin{subfigure}{.45\linewidth}
       \centering
        \includegraphics[width=0.95\linewidth]{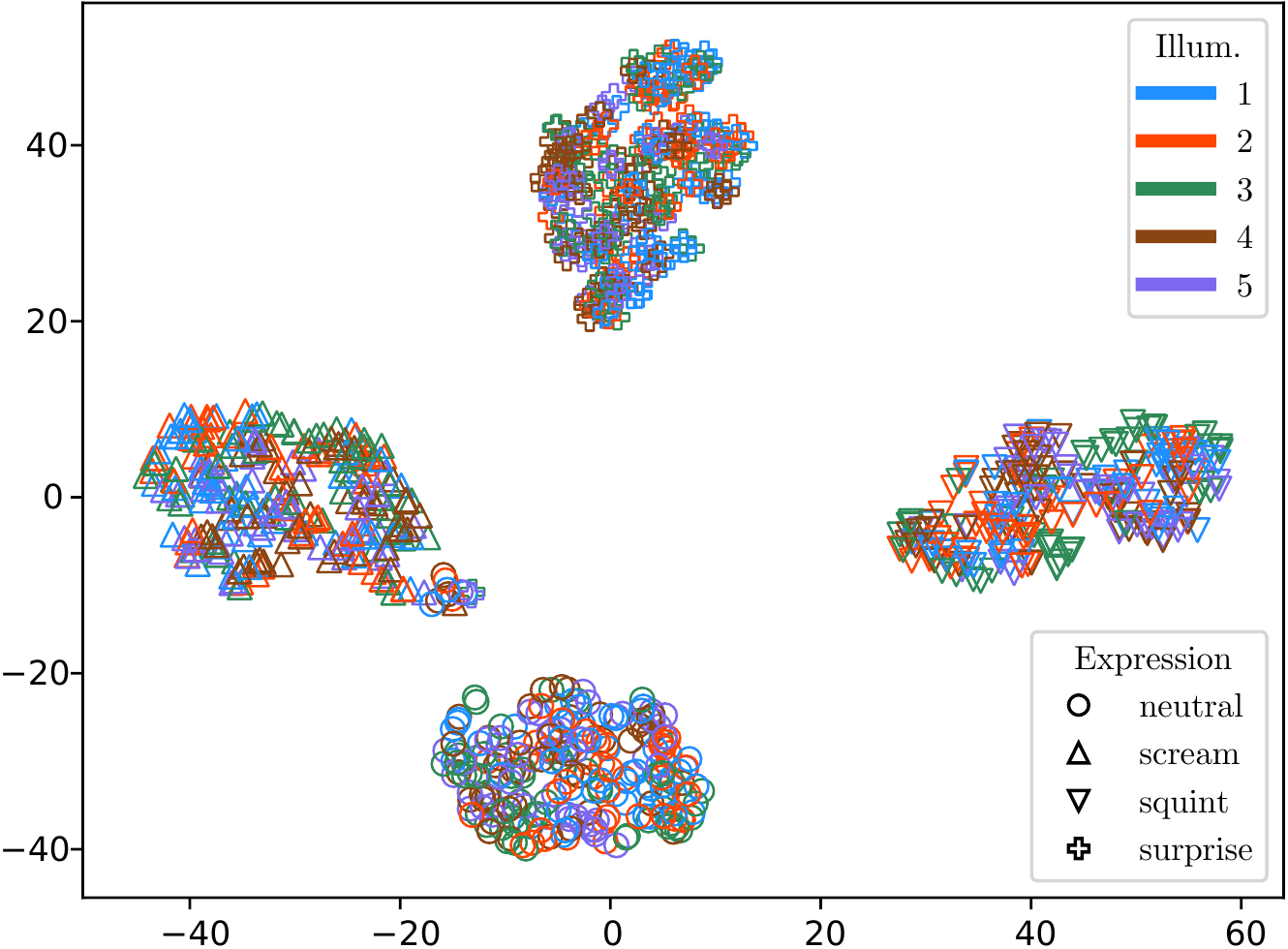}
        \caption{}
        \label{fig:dim-with}
        \qquad
    \end{subfigure}
    \caption{Ablation Study I: Visualising the t-SNE embeddings of the ``expression'' encodings, when training a shallow encoder using (a) only the latent classification loss (Eq. \ref{eq:z-cls}), and (b) using both the classification and disentanglement losses (Eq. \ref{eq:ambiguity-objective}).   Note that {\it without} using the disentanglement loss (a), the variance of the ``illumination'' attribute is clearly captured in the structure of the ``expression'' embeddings.  This is in contrast to using the full proposed loss in (b), where information regarding the illumination attribute is killed and the distribution of illumination labels becomes uniform within each expression cluster. }
    \label{fig:multi-encodings}
\end{figure*}
\subsection{Model Exploration}
\label{sec:modelexploration}
\looseness-1In this section, we present a set of exploratory experiments to verify the properties emerging from the design of the proposed model, and highlight the effectiveness of the proposed loss function.  

\subsubsection{Ablation Study I: Dimensionality Reduction}
\looseness-1Firstly, we present an ablation study on the MultiPIE database, considering particularly the ``expression'' and ``illumination'' attributes.  In this study, we train only  $M$ (shallow, fully-connected) attribute encoders and complimentary classifiers, with each corresponding to the expression and illumination attributes.  Furthermore, we optimize this network with respect to the disentanglement $\mathcal{L}_{dis}$ and classification losses $\mathcal{L}^{\mathbf{x}}_{cls}$ (Eq. \ref{eq:ambiguity-objective}, \ref{eq:z-cls}), that is 
\begin{equation}
\mathcal{L}_{abl}=\mathcal{L}^{\mathbf{x}}_{cls}+\lambda_{dis}\mathcal{L}_{dis}
\end{equation}
and compare the resulting embeddings to the same network trained on just a classification loss (i.e., with $\lambda_{dis}=0$).   

Results for both scenarios are presented in \cref{fig:multi-encodings}, where the t-SNE embeddings semantically corresponding to the ``expression'' label are visualized and compared.  In both cases, the proposed classifier design successfully instils label information in the resulting embeddings, empirically validating the utility of the proposed $\mathcal{L}_{cls}^\mathbf{x}$ term for generating low-dimensional representations of the attributes, as clearly shown by the clustering results.  However, without including the additional disentanglement loss (\cref{fig:dim-without}) the embeddings  carry useful information regarding the ``illumination'' attribute, clearly reflected in the structure of each expression cluster.    On the contrary, when including the proposed disentanglement loss (\cref{fig:dim-with}), variance due to the ``illumination'' attribute is killed--leading to an effectively uniform distribution across illumination labels within each ``expression'' cluster.  To provide further quantative evidence, we compute the Hopkins Statistic \cite{hopkins} as employed in \cite{validating-clusters} in order to validate cluster tendency within each expression cluster in each experiment visualized in \cref{fig:multi-encodings}.  In more detail, we do expect some variation existing in the clusters due to expression-intensity variation. However, illumination variation should be removed under the proposed disentanglement loss.  This is verified by the quantitative results presented in \cref{table:hopkins}.  Specifically, results using the proposed loss show a very low cluster tendency, since values close to $0.5$ indicate randomly distributed data drawn from a uniform distribution.  This is in contrast to results without the proposed loss, with values closer to $1$.  We note that values greater than $0.75$ indicate a clustering tendency at the $90\%$ confidence level.
This further validates  the effectiveness of the composite loss function proposed in this work, and suggests these loss terms are also suitable for general purpose dimensionality reduction tasks.
\begin{table}[]
\begin{tabular}{lll}
\cline{2-3}
\multicolumn{1}{l|}{}                    & \multicolumn{2}{c|}{Mean Hopkins Statistic (per illumination label)}                                  \\ \hline
\multicolumn{1}{|c|}{Expression Cluster} & \multicolumn{1}{c|}{Disentangle Loss ($\lambda_{dis}=1$)}                         & \multicolumn{1}{c|}{Ablation ($\lambda_{dis}=0$)}                     \\ \hline
\multicolumn{1}{|c|}{neutral}            & \multicolumn{1}{c|}{$\mathbf{0.62 \pm 0.10}$} & \multicolumn{1}{c|}{$0.80 \pm 0.11$} \\ \hline
\multicolumn{1}{|c|}{scream}             & \multicolumn{1}{c|}{$\mathbf{0.52 \pm 0.11}$} & \multicolumn{1}{c|}{$0.86 \pm 0.08$} \\ \hline
\multicolumn{1}{|c|}{squint}             & \multicolumn{1}{c|}{$\mathbf{0.57 \pm 0.08}$} & \multicolumn{1}{c|}{$0.73 \pm 0.17$} \\ \hline
\multicolumn{1}{|c|}{surprise}   & \multicolumn{1}{c|}{$\mathbf{0.59 \pm 0.10}$}  & \multicolumn{1}{c|}{$0.74 \pm 0.07$}                     \\ \hline
\end{tabular}
\caption{The Hopkins Statistic computed for each expression cluster separately: values close to $1.0$ indicate heavily clustered data, whereas values near $0.5$ are indicative of randomly distributed data.}
\label{table:hopkins}
\end{table}

\begin{figure}
    \centering
    \includegraphics[width=1.0\linewidth]{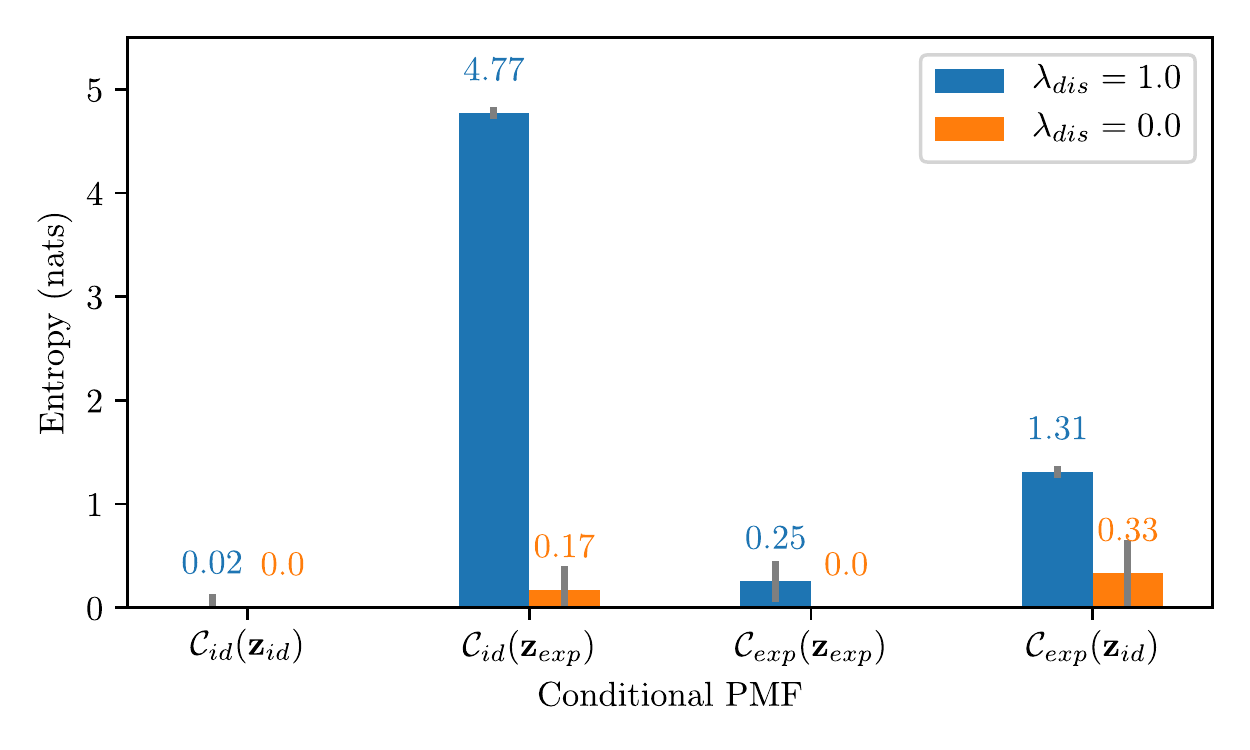}
    \caption{
Classifier distribution entropy over expression and identity classes in MultiPIE, evaluated with ($\lambda_{dis}=1.0$) and without ($\lambda_{dis}=0.0$) the proposed disentanglement loss.  A high-entropy distribution is successfully induced indicating disentanglement, while entropy is near-zero when evaluated on matching attribute encodings.
    \label{fig:cls_cond_entropy}}
\end{figure}

\begin{figure}
    \centering
    \includegraphics[width=\linewidth]{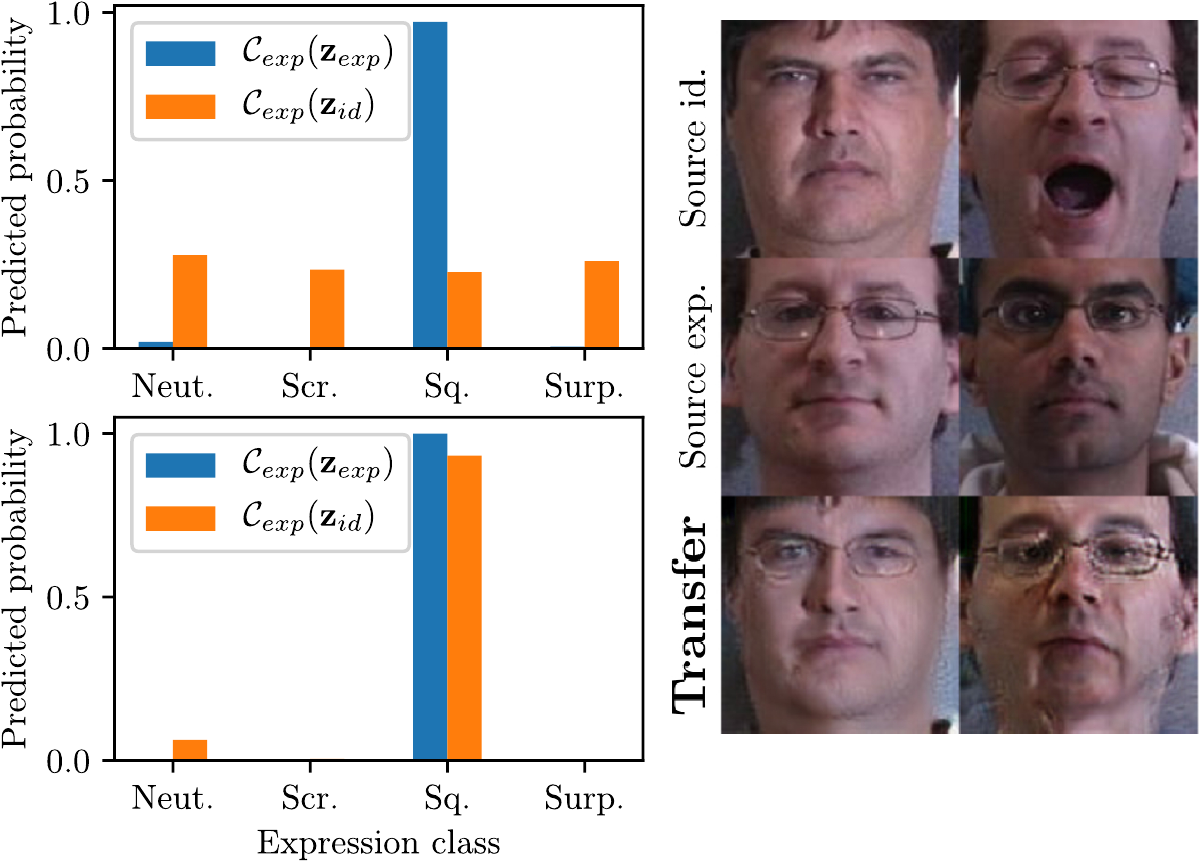}
    \caption{\textbf{Left}: Predicted classes (top row: with disentanglement, bottom row: without disentanglement loss).  
    Right:  Ablation study showing that without the proposed disentanglement loss, identity content is transferred in the results.\label{fig:modelexplclassconf}
    }
\end{figure}

\subsubsection{Ablation Study II: Model Validation}
In this section, we present a further ablation study by utilizing the full proposed model (\cref{sec:meth}) trained on the MultiPIE database.  In particular, for this study we compare the full model to a limited version that does {\it not} include the disentanglement loss (\cref{eq:ambiguity-objective}) ($\lambda_{dis}=0$).  For validation, we firstly compute the entropy of the class conditional Probability Mass Functions (PMFs) of classifiers for expression ($C_{exp}$) and identity ($C_{id}$), when applied on expression ($\mathbf{z}_{exp}$) and identity  ($\mathbf{z}_{id}$) encodings respectively, visualized in \cref{fig:cls_cond_entropy}.  As can be clearly seen, the entropy of the classifier PMF for $C_{m}(\mathbf{z}_{m'})$ with $m\neq m'$ increases substantially, as expected based on the proposed model design, with no significant changes otherwise.  Note that the difference in entropy magnitude for the {\it identity} and {\it expression} classifiers is due to the vastly larger number of classes for identity (147) compared to expression (4).  To further validate these conclusions, in \cref{fig:modelexplclassconf} (first column) we visualize the conditional PMFs of classifiers for ``expression'' and ``identity'' when applied on both attribute encodings derived by the model.  As can be clearly seen, without explicitly optimizing for the proposed disentanglement loss, the PMF clearly indicates that information regarding ``expression'' still remains in the ``identity'' embeddings, that can successfully predict the expression of the subject (first column, bottom row).  By including the proposed loss, the class-conditional probabilities for each individual expression are no longer present in the ``identity'' embeddings (first column, top row)--as the probability mass is spread uniformly across all individual expressions.  This further confirms our hypothesis, that the proposed loss function preserves crucial discriminative information for attributes-of-interest, while at the same time kills any variance related to other attributes (in this case, expression). Finally, the utility of this loss term is further evinced qualitatively by an ablation study, showing that without the disentanglement losses, identity components and ghosting artefacts from clothing are prone to mistakenly fall into the expression representations \cref{fig:modelexplclassconf} (second column).

\subsection{Expression Synthesis}
\label{sec:experiments:synthesis}
In this section, we present a set of {\it expression synthesis experiments} performed across several databases.  Note that for most GAN-based methods, synthesis is performed by conditioning  on a specific target expression, usually in the form of a binary label (e.g., ``neutral'' to ``smile'').  This is in contrast to the proposed method that is able to capture intra-class variability, and is therefore capable of generating varying intensity and style images of the same target expression.  We can also obtain a representation equivalent to an expression label as used in other models (such as \cite{starGAN}) by simply taking the expected value of  the embeddings, $\mathbb{E}_{i=1}^{N}[\mathbf{z}^{(i)}_m]$. 

\begin{figure*}
    \centering
    \includegraphics[width=1.0\linewidth]{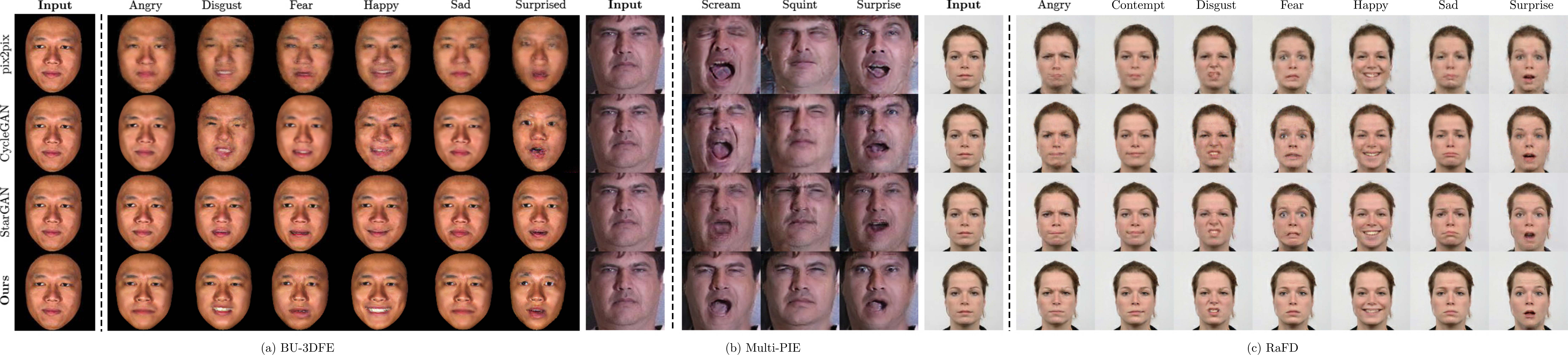}
    \caption{Expression synthesis comparison against baseline methods, on the test set of all datasets (zoom for quality).}
    \label{fig:holdout}
\end{figure*}

We compare our method against SOTA image-to-image translation models, applied on the test sets of BU, MultiPIE and RaFD in Fig. \ref{fig:holdout} (a), (b), and (c) respectively.  As can be seen, the proposed method can generate sharp, realistic images of target expression, while capturing expression intensity.   To provide further evidence on the expression synthesis task, we present a set of quantitative experiments on expression recognition.  Specifically, we utilize all test set images with neutral expressions across all databases, following the data-splits described in \cref{sec:datasets}.  Subsequently, we translate to all expressions for all compared methods, and train a simple CNN expression classifier on the corresponding training sets for each dataset.   Evaluation is performed by utilizing both the classification accuracy on the test set, along with the Fr\'{e}chet Inception Distance (FID)~\cite{fid}--considered as an improvement over the Inception Score for capturing the similarity of generated images to real images.  As can be seen in ~\cref{fig:test-quant}, the proposed method outperforms compared techniques on nearly all databases and metrics.  Note that only the proposed method and StarGAN \cite{starGAN} can accommodate multi-attribute/domain transfer, and therefore one network is trained for all expressions.  This does not hold for methods such as CycleGAN and pix2pix, where a separate instance of the network needs to be trained for each expression pair.   Although in theory this could provide an advantage to the methods trained pair-wise as each network is tailored to a specific transfer task, in practice the proposed method achieves better results than ``paired'' methods.

\begin{table}
\setlength{\tabcolsep}{1.5pt}
  \scriptsize
  \centering
    \begin{tabular}{|r|r|r|r|r|r|r|r|r|r|}
    \hline
          & \multicolumn{3}{c|}{\textbf{MultiPIE}}
          & \multicolumn{3}{c|}{\textbf{BU}}
          & \multicolumn{3}{c|}{\textbf{RaFD}} \\
          \hline
           \textbf{Model} &
           \multicolumn{1}{l|}{\textbf{mean}} & \multicolumn{1}{c|}{\textbf{std}} & \multicolumn{1}{c|}{\textbf{FID}} &
           \multicolumn{1}{l|}{\textbf{mean}} & \multicolumn{1}{c|}{\textbf{std}} & \multicolumn{1}{c|}{\textbf{FID}} &
           \multicolumn{1}{l|}{\textbf{mean}} & \multicolumn{1}{c|}{\textbf{std}} & \multicolumn{1}{c|}{\textbf{FID}}
           \\
                    \hline
    pix2pix  & $0.99$ &  $\pm 0.01$ & $121.86$ & $\mathbf{0.89}$ & $\mathbf{\pm0.02}$ & $58.99$ & $\mathbf{0.99}$ &  $\mathbf{\pm0.01}$ & $91.57$ \\
    CycleGAN & $0.96$ &  $\pm 0.01$ & $97.12$ & $0.63$  & $\pm0.02$ & $65.55$ & $0.93$ &  $\pm0.01$ & $71.44$ \\
    StarGAN  & $0.99$  &  $\pm0.01$  & $\mathbf{75.39}$ & $0.74$  & $\pm0.03$ & $59.58$ & $0.95$ & $\pm0.02$ & $55.99$ \\
    \textbf{Ours} & $\mathbf{1.0}$  & $\mathbf{\pm0.00}$ & $90.95$ & $0.77$ & $\pm0.02$ & $\mathbf{46.62}$ &
    $0.94$ & $\pm0.01$ & $\mathbf{52.44}$ \\
    \hline
    \end{tabular}%
    \caption{Classification accuracy (mean, std) and FID\cite{fid} for expression syntheses on the test set. Whilst pix2pix performs well under classification, it performs poorly under the FID metric, and the samples are noticeably less sharp.}
  \label{fig:test-quant}
\end{table}%
\begin{figure*}
    \centering
    \includegraphics[width=1.0\linewidth]{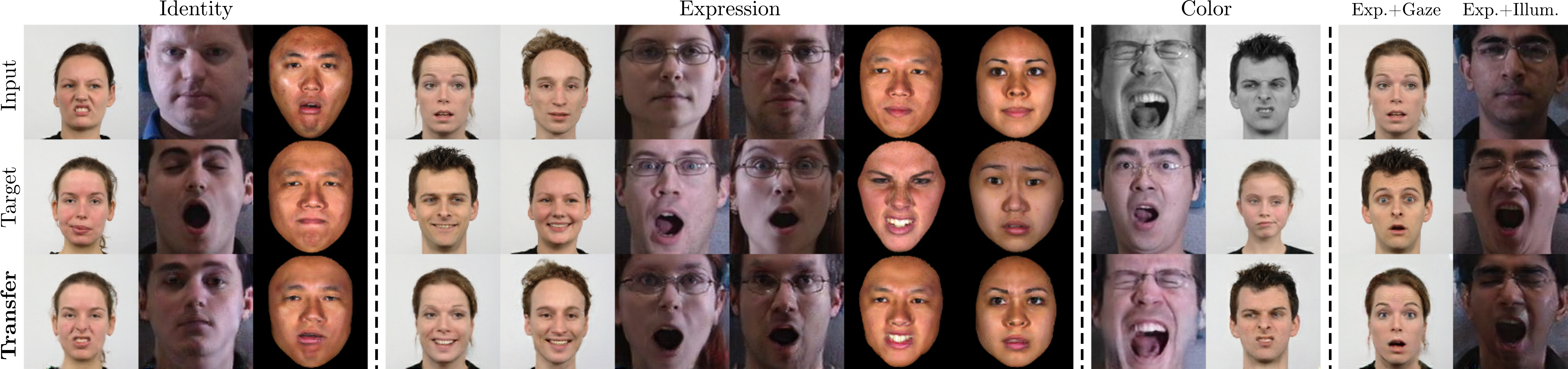}
    \caption{Multiple-attribute transfer across  databases and attributes. \textbf{Row 1}: Input image; \textbf{Row 2}: Target image; \textbf{Row 3}: Single or Joint attribute transfer (zoom in for better quality). }
    \label{fig:all-transfers-experimenalsec}
\end{figure*}
\subsection{Multi-Attribute Transfer}
\label{sec:experiments:atttransfer}
\looseness-1In this section, we present experiments that involve arbitrary, intensity-preserving transfer of attributes.  While most other methods require a target domain or a label, in our case we can simply swap the obtained representations arbitrarily from source sample to target sample, while also being able to perform arbitrary operations on the embeddings--without requiring labels for the transfer.  In more detail, in~\cref{fig:all-transfers-experimenalsec}, we demonstrate the multi-attribute transfer capabilities of the proposed method.  By mapping from the image space to a semantically decomposed latent structure with generalizable properties, we can successfully transfer intrinsic facial attributes such as identity, expression, and gaze, as well as appearance-based attributes such as illumination and image color.
 Note that this demonstrates the generality of our method, handling arbitrary sources of variation.  Finally, we also show that it is entirely possible to transfer several attributes jointly, by using the corresponding representations.  We show examples where we simultaneously transfer expression and illumination, as well as expression and gaze.
\subsection{Attribute Mixtures and Interpolation}
\label{sec:experiments:novelidinter}
\begin{figure}[h]
    \centering
    \includegraphics[width=0.9\linewidth]{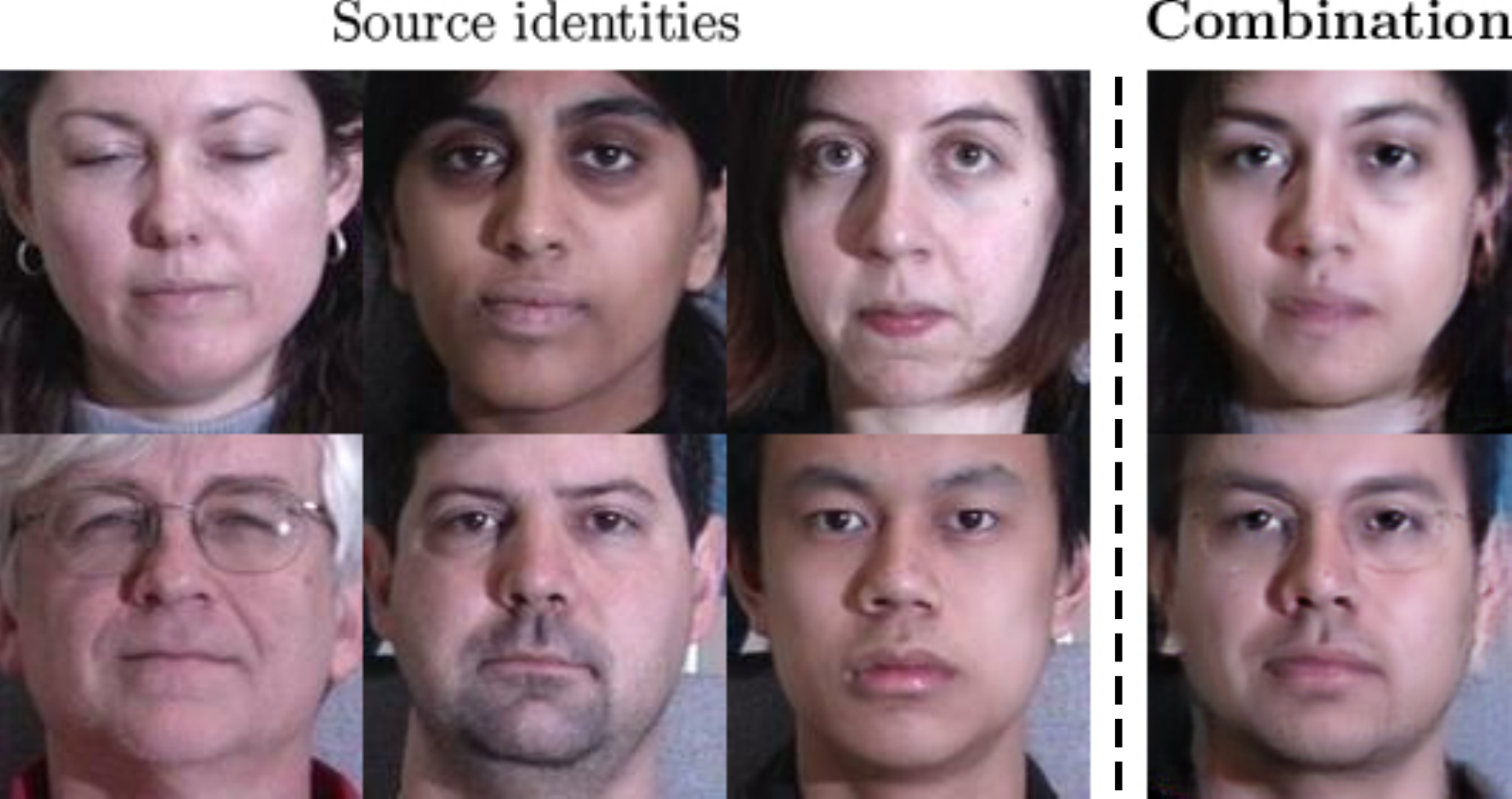}
    \caption{Combining multiple identity representations.}
    \label{fig:convex-id}
\end{figure}
\begin{figure*}[ht]
    \centering
    \includegraphics[width=0.9\linewidth]{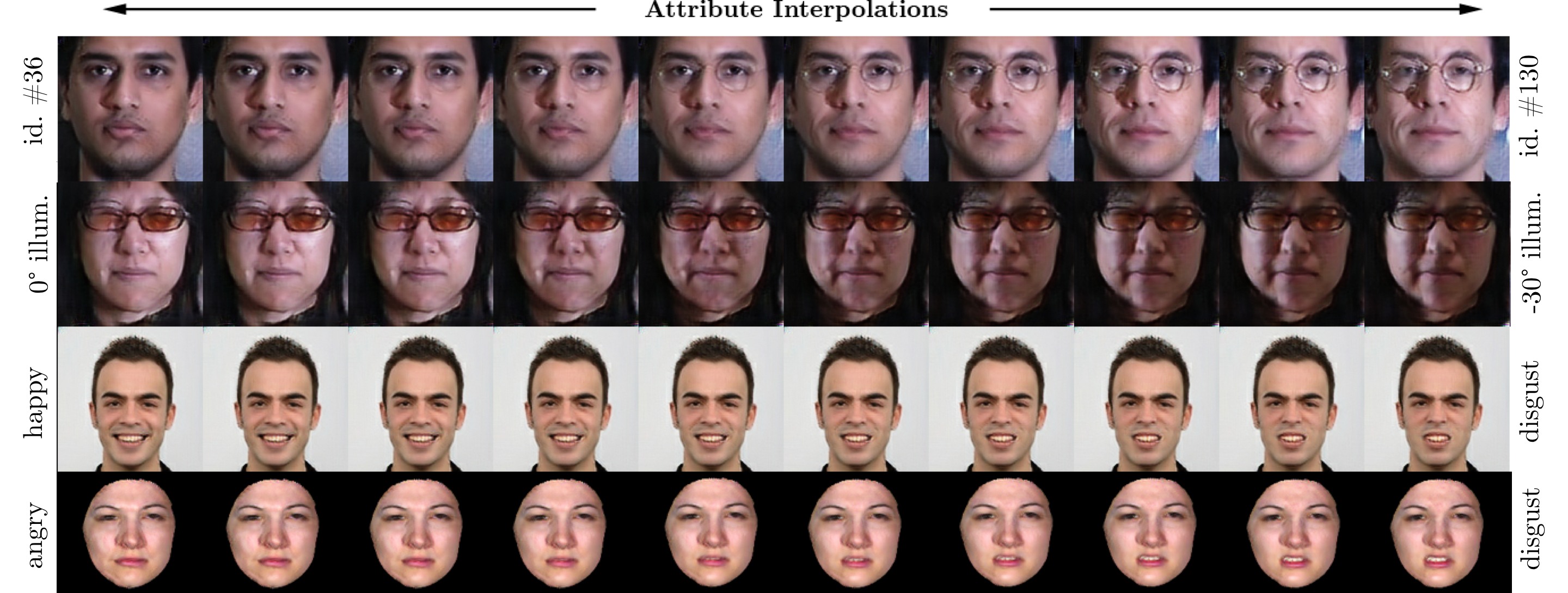}
    \caption{Linearly interpolating between various categorical target attributes (identity, illumination, expression).}
    \label{fig:interpolation}
\end{figure*}
\begin{figure*}
    \centering
    \includegraphics[width=1.0\linewidth]{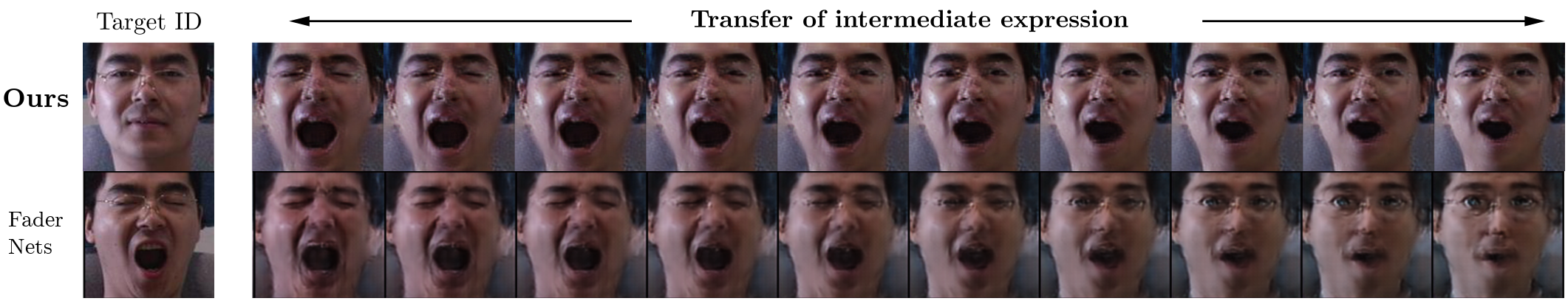}
    \caption{Interpolating  expression representations between two images with labels: `scream' and `surprise'.  Results with the proposed method ({row $1$})  transfer the interpolated representation onto an arbitrary unseen test identity, in an entirely different expression.  This is compared with FaderNets (row $2$).}
    \label{fig:fader-compare-interp}
\end{figure*}
\begin{figure*}
    \centering
    \includegraphics[width=1.0\linewidth]{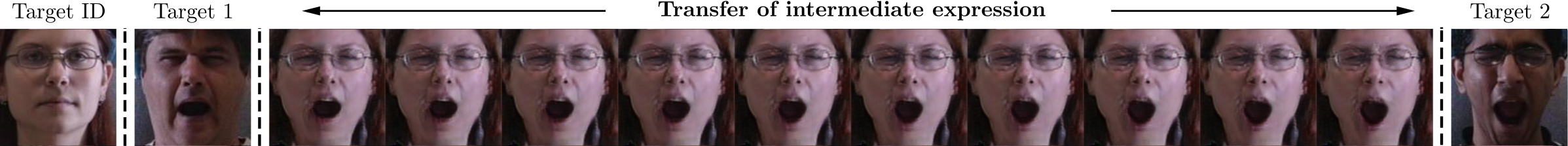}
    \caption{Our method affords the ability to interpolate expression representations of the {\it same} label, highlighting the extent to which intra-attribute variance is preserved. Pictured here is the result of interpolating the expression representations of two {\it unseen} images with varying intensity expressions (``scream''), while transferring the interpolated expression onto an also {\it unseen} target identity.   
}
    \label{fig:interp-same-label}
\end{figure*}
Our method produces generalizable representations that can be combined in many ways to synthesize novel images.  This is an important feature that could be used towards tasks such as data augmentation, as well as enhancing the robustness of classifiers for face recognition.  In this section, we present experiments to demonstrate that the learned representations are both {\it generalizable} and {\it continuous} in the latent domain for all attributes.  Consequently, latent contents can be manipulated accordingly in order to synthesize entirely novel imagery. Compared to dedicated methods for interpolation employing GANs such as~\cite{fader}, our method readily allows for categorical attributes and also preserves intensity when interpolating between specific values.  In more detail, in~\cref{fig:convex-id} we decode the convex combination of $3$ identity embeddings from distinct samples ($i$, $j$, $k$), and synthesize a realistic mixture of the given identities rendered as a new person by pushing the resulting representations $\mathbf{z'}$ through the trained decoder $\mathcal{D}$, where
\begin{equation}
{\mathbf{{z'}}=\alpha_1\mathbf{z}^{(i)}_{id}+\alpha_2\mathbf{z}^{(j)}_{id}+\alpha_3\mathbf{z}^{(k)}_{id}, \sum \alpha_i =1, \alpha_i>0}.
\end{equation}
As can be seen, the resulting identities bear characteristics from all three source identities, while generating visually distinct identities.  

To further demonstrate the {\it generalizable} and {\it continuous} nature of the resulting embeddings, we perform a set of experiments focused on latent interpolation.  Given embeddings derived by the proposed method for a given attribute $m$ and two distinct data samples $\{i,j\}$, we linearly interpolate between the latent representations, that is,
\begin{equation}
\mathcal{\hat{Z}}_m = \left\{\, \alpha \mathbf{z}_m^{(i)} + \left(1-\alpha\right) \mathbf{z}_m^{(j)} \bigm| 0\le \alpha\le 1 \,\right\},
\end{equation}
where each element of $\mathcal{\hat{Z}}_m$ is pushed through the decoder $\mathcal{D}$ to generate the corresponding images.  

\looseness-1In~\cref{fig:interpolation}, we show latent interpolation results between various categorical attributes, interpolating between different identities (row $1$), different illumination conditions (row $2$), as well as different expressions (rows $3$ and $4$).  As can be observed, the resulting images are both realistic, as well as impressively clear and sharp given the challenging setting--that is, the network has {\it never} observed transitions between expressions, but can readily synthesize realistic images that describe this transition, that are smooth enough to generate realistic videos of expression transitions.  

\looseness-1In~\cref{fig:fader-compare-interp}, we show the result of an experiment that includes an unseen target identity, and two randomly selected images from the target expressions ``scream'' and ``surprise''.  We extract the identity embeddings from the unseen image using our method, and subsequently replace them in the two sample images that display the two expressions (leftmost and rightmost decoded images in row $1$).  Subsequently, we interpolate the expression encodings between the two different expressions, showing intermediate results.  As can be seen, the proposed method can easily handle the varying intensity expressions jointly with identity transfer and can produce realistic results throughout the interpolation steps.  In row $2$, we present results from applying the same experiment with FaderNets.  Clearly, FaderNets have difficulties in retaining the input image's identity during transfer, with the interpolation steps lacking smoothness and sharpness. Finally, in \cref{fig:interp-same-label}, we present another challenging setting where we interpolate expressions {\it within} the same expression attribute label.  That is, given a target identity, and two different images that portray the same emotional expression (in this case, ``scream''), we interpolate between the expression embeddings while at the same time transfer the target identity onto the resulting images.  While this experiment serves to further demonstrate the multitude of ways by which we can flexibly manipulate the latent embeddings and produce meaningful results, it is also important to note that methods relying on binary attributes (e.g. StarGAN) are by construction unable to interpolate between expressions of the same class label.

\section{Conclusion}
\looseness-1In this paper, a novel method for learning disentangled and generalizable representations of visual attributes has been presented.  The intuition behind the proposed technique lies in utilizing a simple objective for ensuring that representations contain discriminative, high-level information with respect to attributes of interest, while ensuring that the representations are invariant to the visual variability sourced by other attributes.  To encourage generalization, the network is trained in such a way that representations capturing attribute variation can be arbitrarily shuffled and combined across samples, giving rise to novel hybrid imagery when simply pushed through a decoder.  The proposed method offers several advantages in comparison to related methods, such as intensity-preserving image translation, multi-attribute transfer, as well as the generation of diverse images with several variations of expressions readily generated on demand.  With several experiments on popular databases such as MultiPIE, BU, and RaFD, we provided experimental evidence to both provide sufficient model exploration to justify the properties of the method, as well as showcase the possibilities arising with the proposed formulation that facilitates attaching attribute semantics to each component of  the latent space, leading to a continuous and generalizable latent structure and the possibility to generate a rich gamut of novel hybrid imagery.

{\small
\bibliographystyle{IEEEtran}

\bibliography{IEEEabrv,egbib}
}

\ifCLASSOPTIONcaptionsoff
  \newpage
\fi

\end{document}